\documentclass{ecai2006}

\usepackage{local}

\title{Adaptation Knowledge Discovery from a Case Base}
\author{M. d'Aquin\thanks{LORIA (UMR 7503 CNRS--INPL--INRIA-Nancy~2--UHP), BP 239, 54506 Vand{\oe}uvre-l{\`es}-Nancy, FRANCE}
        \and
       F. Badra$^1$
       \and
       S. Lafrogne$^1$
       \and
       J. Lieber$^1$
       \and
       A. Napoli$^1$
       \and
       L. Szathmary$^1$}
\authorfoot{M. d'Aquin \and F. Badra \and S. Lafrogne
       \and J. Lieber \and A. Napoli \and L. Szathmary}
\begin{document}
%%%%%%%%%%%%%%%%

\maketitle

\bibliographystyle{ecai2006}

\begin{abstract}
In case-based reasoning,
 the adaptation step
%==AVPVC==%  solve the target problem is
%==AVPVC==% %==AVPVC==%  at the same time crucial and
%==AVPVC==%  difficult to implement
%==AVPVC==%  because
%==AVPVC==% %==AVPVC==%  The reason for this difficulty is that,
%==AVPVC==%  in general, adaptation strongly
 depends in general on domain-dependent knowledge,
 which motivates studies on adaptation knowledge acquisition
 (\aca).
\cabamaka is an \aca system
 based on principles of knowledge discovery from
 databases.
%==AVPVC==%  and data-mining.
%==AVPVC==% It is implemented in \cabamaka,
This system explores the variations within
 the case base to elicit adaptation knowledge.
It has been successfully tested in an application of
 case-based
%==AVPVC==%  reasoning to
 decision support to breast
 cancer treatment.
\end{abstract}

%        ------------
\section{INTRODUCTION}
%        ------------
\label{sec:introduction}

Case-based reasoning (\rapc~\cite{Riesbeck-Schank89})
 aims at solving a target problem thanks
 to a case base.
A case represents a previously solved problem.
%==AVPVC==%  and may be seen as a pair $(\text{problem}, \text{solution})$.
A \rapc system selects a case from the case base
 and then
 adapts the associated solution, requiring domain-dependent knowledge
 for adaptation.
The goal of adaptation knowledge acquisition (\aca)
 is to
%==AVPVC==%  detect and
 extract this knowledge.
%==AVPVC==% This is the function of
The system \cabamaka
 applies principles of knowledge discovery from databases (\ecbd)
 to \aca.
%==AVPVC==%  in particular frequent itemset extraction.
%==AVPVC==% This paper presents this system.
%==AVPVC==%  its principles, its implementation and
%==AVPVC==%  an example of adaptation rule discovered %%**S'IL RESTE !!!
%==AVPVC==%  in the framework of an application
%==AVPVC==%  % of \rapc
%==AVPVC==%  to breast cancer treatment.
The originality of \cabamaka lies essentially in the approach to \aca
 that uses a powerful learning technique that is guided by a domain
 expert,
% assisted by a computer scientist,
 according to the spirit of \ecbd.
This paper proposes an original and working approach to \aca,
 based on \ecbd techniques.
%==AVPVC==% This is one of the rare papers trying to build an effective
%==AVPVC==%  bridge between knowledge discovery and case-based reasoning.

%==AVPVC==% The paper is organized as follows.
%==AVPVC==% Section~\ref{sec:rapc-adaptation} presents basic notions
%==AVPVC==%  about \rapc and adaptation.
%==AVPVC==% Section~\ref{sec:travaux-aca} summaries researches on \aca.
%==AVPVC==% Section~\ref{sec:cabamaka} describes the system \cabamaka:
%==AVPVC==%  its main principles, its implementation and examples of
%==AVPVC==%  adaptation knowledge acquired from it.
%==AVPVC==% Finally, section~\ref{sec:conclusion} draws some conclusions
%==AVPVC==%  and points out the future work.

%          ---------------------
\paragraph{\rapc and adaptation.}
%          ---------------------
%
A case in a given \rapc application
% encodes a problem-solving episode
 is usually represented by a
 pair $(\prb, \sol(\prb))$ where
 $\prb$ represents a
 problem statement and
 $\sol(\prb)$, a solution of $\prb$.
%Thus, it represents at least
% a \emph{problem} $\prb$ and
% a \emph{solution} $\sol(\prb)$ of $\prb$.
%It is denoted by the pair $(\prb, \sol(\prb))$.
%==AVPVC==% Let $\Problemes$ and $\Solutions$
%==AVPVC==%  be the set of problems and the set of solutions of the application
%==AVPVC==%  domain,
%==AVPVC==%  and ``is a solution of'' be a binary relation on
%==AVPVC==%  $\Problemes\times\Solutions$.
%==AVPVC==% In general, this relation
%==AVPVC==% % may be incompletely known,
%==AVPVC==%  is not known in the whole
%==AVPVC==%  but at least a finite
%==AVPVC==%  number of its instances
%==AVPVC==%  $(\prb, \sol(\prb))$ is known and constitutes the case base
%==AVPVC==%  $\basedecas$.
\rapc relies on
 the \emph{source cases} $(\source, \sol(\source))$
 that constitute the \emph{case base} $\basedecas$.
%==AVPVC==% An element of $\basedecas$ is called a \emph{source case}
%==AVPVC==%  and is denoted by $\cassource=(\source, \sol(\source))$,
%==AVPVC==%  where $\source$ is a \emph{source problem}.
In a particular \rapc session, the problem to be solved
 is called \emph{target problem}, denoted by $\cible$.
A case-based inference associates to $\cible$ a solution $\sol(\cible)$,
% thanks to
 with respect to
 the case base $\basedecas$ and to
 additional knowledge bases, in particular $\Ontologie$, the
 \emph{domain ontology}
%==AVPVC==%  (also known as domain theory or domain knowledge)
 that usually introduces the concepts and terms
 used to represent the cases.
%==AVPVC==% It can be noticed that the current research work makes the
%==AVPVC==%  assumption that there exists a domain ontology associated with
%==AVPVC==%  the case base,
%==AVPVC==%  in the spirit of knowledge-intensive \rapc~\cite{aamodt90}.

A classical decomposition of \rapc consists in
 the steps of retrieval and adaptation.
\emph{Retrieval} selects
 $(\source, \sol(\source))\in\basedecas$ such that $\source$
 is judged to be similar to $\cible$.
%==AVPVC==%  according to some similarity criterion.
The goal of adaptation is to solve $\cible$ by modifying
 $\sol(\source)$ accordingly.
%==AVPVC==% Thus, the profile of the adaptation function is
%==AVPVC==% \begin{equation*}
%==AVPVC==%   \fm{Adaptation} :
%==AVPVC==%   ((\source, \sol(\source)), \cible)
%==AVPVC==%   \mapsto
%==AVPVC==%   \sol(\cible)
%==AVPVC==% \end{equation*}

The work presented hereafter is based on the following model of
 adaptation, similar to
 \emph{transformational analogy}~\cite{carbonell83}:
\begin{dingautolist}{192}
\item\label{etape:appariement}
%  From $\source$ and $\cible$,
%   $\diffPb$, that encodes the similarities and dissimilarities
%   of these two problems, is built:
%   $(\source, \cible)\mapsto\diffPb$.
  $(\source, \cible)\mapsto\diffPb$,
   where $\diffPb$ encodes the similarities and dissimilarities
   of the problems $\source$ and $\cible$.
\item\label{etape:parallele}
%  From $\diffPb$ \emph{and} adaptation knowledge $\CA$,
%   $\diffSol$, that encodes the similarities and dissimilarities of
%   $\sol(\source)$ and (what will be) $\sol(\cible)$, is computed:
%   $(\diffPb, \CA)\mapsto\diffSol$.
  $(\diffPb, \CA)\mapsto\diffSol$,
  where $\CA$ is the adaptation knowledge
  and
  where $\diffSol$ encodes the similarities and dissimilarities of
  $\sol(\source)$ and the forthcoming $\sol(\cible)$.
\item\label{etape:modification}
%  Finally, $\sol(\source)$ is modified into $\sol(\cible)$ according
%   to $\diffSol$:
%   $(\sol(\source), \diffSol)\mapsto\sol(\cible)$.
  $(\sol(\source), \diffSol)\mapsto\sol(\cible)$,
  $\sol(\source)$ is modified into $\sol(\cible)$ according
   to $\diffSol$.
\end{dingautolist}

Adaptation is generally supposed to be
 domain-dependent in the sense that it relies on
 domain-specific adaptation knowledge.
Therefore, this knowledge has to be acquired.
This is the purpose of \emph{adaptation knowledge acquisition}
 (\aca).

%          -----------------------
\paragraph{A related work in \aca.}
%          -----------------------
\label{sec:travaux-aca}
The idea of the research presented in~\cite{hanney96} %,hanney97}
 is to exploit the variations between source cases to learn
 adaptation rules.
These rules compute variations on solutions from variations
 on problems.
% that represent how variations on solutions can be
% computed from variations on problems.
More precisely, ordered pairs $(\cassourceN1, \cassourceN2)$
 of similar source cases are formed.
Then, for each of these pairs,
 the variations between the problems $\src1$ and $\src2$
 and the solutions $\sol(\src1)$ and $\sol(\src2)$
 are represented
 ($\diffPb$ and $\diffSol$).
% ($\diffPb$ and $\diffSol$ in the notations of this paper).
Finally, the adaptation rules are learned, using as training
 set the set of the input-output pairs $(\diffPb, \diffSol)$.
%==AVPVC==% This approach has been tested in two domains:
%==AVPVC==%  the estimation of the price of flats and houses %properties
%==AVPVC==%  and the prediction of the rise time of a servo mechanism.
The experiments have shown that the \rapc system using
 the adaptation knowledge acquired from the automatic system
 of \aca
 shows a better performance compared to the \rapc system
 working without adaptation.
% gives some improvement compared to the same
% \rapc system performing a null adaptation.
This research has strongly influenced our work
 that is globally based on similar ideas.
% which is based
% on similar ideas, with some important differences, as the next
% section shows.

%==AVPVC==% %\remJeanEnTexte{Petit paragraphe pour dire ce que ces différents
%==AVPVC==% %  travaux partagent, en particulier, travail à partir des
%==AVPVC==% %  couples de cas sources (est-ce vrai pour Leake ? Pour Anand ?)}
%==AVPVC==% %\LignesPoints{4}
%==AVPVC==% %These approaches to \aca %with the exception of ???
%==AVPVC==%  shares the idea of exploiting adaptation cases.
%==AVPVC==% For some of them (\cite{jarmulak01,leake96b}), %**VOIR QUEL Leake !!
%==AVPVC==%  the adaptation cases themselves constitute the adaptation
%==AVPVC==%  knowledge and are exploited thanks to a recursive \rapc
%==AVPVC==%  process.
%==AVPVC==% For the other one %ones?
%==AVPVC==%  (\cite{hanney97}),
%==AVPVC==%  as for the approach presented in this paper,
%==AVPVC==%  the adaptation cases are the input of a learning process.

%        ---------
\section{\CABAMAKA}
%        ---------
\label{sec:cabamaka}

%==AVPVC==% We now present the \cabamaka system, for acquiring adaptation
%==AVPVC==%  knowledge.
%Thereafter,
% the so-called \cabamaka system that is
% developed for acquiring adaptation knowledge
% is presented.
%==AVPVC==% The \cabamaka system is at present working in the medical
%==AVPVC==%  domain of cancer treatment,
%==AVPVC==%  but it may be reused in other application domains where there
%==AVPVC==%  exist problems to be solved by a \rapc system.

%\cabamaka is a system developed for the purpose of \aca
% for a \rapc system in the domain of cancer treatment
% and designed to be reusable for other \rapc applications.

%          -----------
\paragraph{Principles.}
%          -----------
%
\cabamaka deals with
 \underline{ca}se \underline{ba}se \underline{m}ining for
 \underline{\aca}.
Although the main ideas underlying \cabamaka are shared
 with those presented in~\cite{hanney96},
 the followings are original ones.
%It is similar in principles to the system of~\cite{hanney96}
% (see section~\ref{sec:travaux-aca}),
% with several differences.
%In particular,
The adaptation knowledge that is mined has to be
 validated by experts and has to be associated with explanations
 that make it understandable by the user.
In this way, \cabamaka may be considered as a semi-automated
 (or interactive) learning system.
Another difference with~\cite{hanney96} lies in the volume
 of the cases that are examined:
 given a case base $\basedecas$ where
 $|\basedecas|=n$,
 the \cabamaka system takes into account every ordered pair
 $(\cassourceN1, \cassourceN2)$
 with $\cassourceN1\neq\cassourceN2$
 (whereas in~\cite{hanney96}, only the pairs of \emph{similar}
  source cases are considered, according to a fixed criterion).
Thus, the \cabamaka system has to cope with
 $n(n-1)$ pairs, a rather large number of elements,
 since in our application $n\simeq750$.
 ($n(n-1)\simeq5\cdot10^5$).
%Another difference is that \cabamaka
% takes \emph{a priori} into account every ordered pair
% $(\cassourceN1, \cassourceN2)$
% with $\cassourceN1\neq\cassourceN2$,
% whereas in~\cite{hanney96}, only the pairs of \emph{similar}
% source cases are considered (according to a fixed criterion).
%A consequence of this is that \cabamaka has to cope with
% $n(n-1)$ pairs, with $n=|\basedecas|$, which is rather large
% for a learning process
% (in our application $n\simeq750$, thus $n(n-1)\simeq5\cdot10^5$).
This is why efficient techniques of
% knowledge discovery from databases (\ecbd~\cite{dunham03,hand01})
 knowledge discovery from databases (\ecbd~\cite{dunham03})
 have been chosen for this system.

%             --------------------
\subparagraph{Principles of \ecbd.}
%             --------------------
%
The goal of \ecbd is to discover knowledge from databases,
 with the supervision of an analyst (expert of the domain).
A \ecbd session usually relies on three main steps:
 data preparation, data-mining and interpretation.
%==AVPVC==%  of the extracted pieces of information.

\emph{Data preparation} is
%==AVPVC==%  mainly
 based on formatting and filtering
 operations.
The formatting operations
%==AVPVC==%  are used to
 transform the data into a
 form allowing the application of the chosen data-mining operations.
The filtering operations are used for removing noisy data and
 for focusing the data-mining operation on special subsets of
 objects and/or attributes.
%\emph{Data preparation} is composed of formatting and filtering
% operations.
%Formatting transforms the data to make them usable by the
% chosen data-mining process.
%The goals of filtering are
% (1) to suppress the data that are suspected to be noisy and
% (2) to influence the current \ecbd session towards specific types
%     of acquired knowledge.

\emph{Data-mining} methods are applied
 % within the data-mining steps for extracting
 to extract
 pieces of information from the data.
These pieces of information have some regular properties allowing
 their extraction.
%\emph{Data-mining} highlights
% pieces of information encoding some regularities in the data.
For example, \charm~\cite{zaki02} is a data-mining algorithm
 that performs efficiently the extraction of
 \emph{frequent closed itemsets}
 (\emph{FCIs}).
\charm inputs a database in the form of a set of transactions,
 each \emph{transaction} $\transaction$
 being a set of boolean properties or \emph{items}.
%==AVPVC==% In other words, the input of \charm is a single table of a
%==AVPVC==%  relational database, where each column's type is boolean.
An \emph{itemset} $\motif$ is a set of items.
The support of $\motif$, $\support(\motif)$, is the proportion of
 transactions $\transaction$ of the database possessing $\motif$
 ($\motif\subseteq\transaction$).
$\motif$ is frequent,
 with respect to a threshold $\seuil\in[0 ; 1]$,
 whenever $\support(\motif)\geq\seuil$.
%$\motif$ is closed if any strict superset $\motif'$ of $\motif$ has a
% strictly lower support,
% i.e. if $\motif'\supseteq\motif$ and
% $\support(\motif')=\support(\motif)$, then $\motif'=\motif$.
$\motif$ is closed if it has no proper superset $\motifJ$
 ($\motif\subsetneq\motifJ$) with the same support.

\emph{Interpretation}
%==AVPVC==% The \emph{interpretation} step
 aims at
 interpretating
%==AVPVC==%  the extracted pieces of information,
 the output of data-mining
 i.e. the FCIs in the present case,
 with the help of an analyst.
In this way, the interpretation step produces new knowledge units
 (e.g. rules).

%==AVPVC==% The \cabamaka system relies on these
%==AVPVC==% % three % 3 ou 4 ? (cf. préparation = formattage + filtrage) Je laisse sans préciser.
%==AVPVC==%  main \ecbd steps as explained below.
%==AVPVC==% % is built on \ecbd's principles and thus it is composed of
%==AVPVC==% % formatting, filtering, mining and interpretation steps.

%             -----------
\subparagraph{Formatting.}
%             -----------
%
The formatting step of \cabamaka inputs the case base $\basedecas$
 and outputs a set of transactions % $\transaction$,
 obtained from the
 pairs $(\cassourceN1, \cassourceN2)$.
It is composed of two substeps.
During the first substep, each
 $\cassource=(\source, \sol(\source))\in\basedecas$ is formatted in
 two sets of boolean properties:
 $\form(\source)$ and $\form(\sol(\source))$.
The computation of $\form(\source)$ consists in translating $\source$
 from the problem representation formalism to $2^{\Proprietes}$,
 $\Proprietes$ being a set of boolean properties.
Possibly, some information may be lost during this
 translation,
%==AVPVC==%  for example, when translating a continuous property into
%==AVPVC==%  a set of boolean properties,
 but this loss has to be minimized.
%This may involve some loss of information about $\source$
% but this loss has to be minimized.
Now, this translation formats an expression $\source$ expressed
 in the framework of the domain ontology $\Ontologie$ to an expression
 $\form(\source)$ that will be manipulated as data,
 i.e. without the use of a reasoning process.
Therefore, in order to minimize the translation loss, it is
 assumed that
 if
 $p\in\form(\source)$
 and
%==AVPVC==% $p\DecouleOntologie{}q$
%==AVPVC==%  $q$ is a consequence of $p$ in the ontology $\Ontologie$
 $p$ entails $q$ (given $\Ontologie$)
 then $q\in\form(\source)$.
%==AVPVC==% \begin{equation}
%==AVPVC==%   \text{if }
%==AVPVC==%   p\in\form(\source)
%==AVPVC==%   \text{ and }
%==AVPVC==%   p\DecouleOntologie{}q
%==AVPVC==%   \quad
%==AVPVC==%   \text{then }
%==AVPVC==%   q\in\form(\source)
%==AVPVC==%   %$q$ must be a property of $\form(\source)$.
%==AVPVC==%   \label{eq:FermetureDeductive}
%==AVPVC==% \end{equation}
%==AVPVC==%  (where $p\DecouleOntologie{}q$ stands for
%==AVPVC==%   ``$q$ is a consequence of $p$'' in the ontology $\Ontologie$).
In other words, $\form(\source)$ is assumed to be deductively
 closed given $\Ontologie$ in the set $\Proprietes$.
The same assumption is made for $\form(\sol(\source))$.
How this first substep of formatting is computed in practice depends
  heavily on the representation formalism of the cases.
% and is presented, for our application, in section~\ref{subsec:implantation}.

The second substep of formatting produces a transaction
 $\transaction=\form((\cassourceN1,\cassourceN2))$
 for each ordered pair of distinct source cases,
 based on the sets of items
 $\form(\src1)$, $\form(\src2)$,
 $\form(\sol(\src1))$ and $\form(\sol(\src2))$.
Following the model of adaptation presented in
 introduction
 (items~\ref{etape:appariement}, \ref{etape:parallele}
  and~\ref{etape:modification}),
 $\transaction$ has to encode the properties of $\diffPb$ and
 $\diffSol$.
$\diffPb$ encodes the similarities and dissimilarities of $\src1$
 and $\src2$, i.e.:
\begin{itemize}
\item
  The properties common to $\src1$ and $\src2$
   (marked by ``$\fm{=}$''),
\item
  The properties of $\src1$ that $\src2$ does not share
%   (marked by ``$\fm{-}$'') and
   (``$\fm{-}$'') and
\item
  The properties of $\src2$ that $\src1$ does not share
%   (marked by ``$\fm{+}$'').
   (``$\fm{+}$'').
\end{itemize}
All these properties are related to problems and thus are marked by
 $\prb$.
$\diffSol$ is computed in a similar way and
 $\form(\transaction)=\diffPb\cup\diffSol$.
For example,
\begin{align}
  &\text{if}
  \quad
  \begin{cases}
    \form(\src1) = \{a, b, c\}
    \quad
    &\form(\sol(\src1)) = \{A, B\} \\
    \form(\src2) = \{b, c, d\}
    \quad
    &\form(\sol(\src2)) = \{B, C\}
  \end{cases}
  \notag\\
  &\text{then}
  \quad
  \transaction = \left\{a\Moins\Pb, b\Eg\Pb, c\Eg\Pb, d\Plus\Pb, A\Moins\Sl, B\Eg\Sl, C\Plus\Sl\right\}
  \label{eq:formatage}
\end{align}
%Vir**
%==AVPVC==% More generally:
%==AVPVC==% \begin{align*}
%==AVPVC==%   \transaction
%==AVPVC==%   &=     \{p\Moins\Pb ~|~ p\in\form(\src1)\backslash\form(\src2)\} \\
%==AVPVC==%   &~\cup \{p\Eg\Pb    ~|~ p\in\form(\src1)\cap\form(\src2)\} \\
%==AVPVC==%   &~\cup \{p\Plus\Pb  ~|~ p\in\form(\src2)\backslash\form(\src1)\} \\
%==AVPVC==%   &~\cup \{p\Moins\Sl ~|~ p\in\form(\sol(\src1))\backslash\form(\sol(\src2))\} \\
%==AVPVC==%   &~\cup \{p\Eg\Sl    ~|~ p\in\form(\sol(\src1))\cap\form(\sol(\src2))\} \\
%==AVPVC==%   &~\cup \{p\Plus\Sl  ~|~ p\in\form(\sol(\src2))\backslash\form(\sol(\src1))\}
%==AVPVC==% \end{align*}

%==AVPVC==% %             ----------
%==AVPVC==% \subparagraph{Filtering.}
%==AVPVC==% %             ----------
%==AVPVC==% %
%==AVPVC==% The filtering operations may take place before, between and after the
%==AVPVC==%  formatting substeps, and also after the mining step.
%==AVPVC==% They are guided by the analyst.

%             -------
\subparagraph{Mining.}
%             -------
%
The extraction of FCIs is computed thanks to
 \charm (in fact, thanks to a tool based on a \charm-like algorithm)
 from the set of transactions.
% $\transaction$. %=\form((\cassourceN1, \cassourceN2))$.
A transaction
 $\transaction=\form((\cassourceN1, \cassourceN2))$ encodes a specific adaptation
 $((\src1, \sol(\src1)), \src2)\mapsto\sol(\src2)$.
An FCI extracted may be considered as a generalization of
 a set of transactions.
%\charm performs a generalization of these specific adaptations:
% an FCI $\motif$
% encodes the generalization of a subset of the
% transactions.
For example, if
 %==AVPVC==% \begin{equation}
 $\motifEx = \left\{a\Moins\Pb, c\Eg\Pb, d\Plus\Pb, A\Moins\Sl, B\Eg\Sl, C\Plus\Sl\right\}$
 %==AVPVC==%   \label{eq:exemple-MFF}
 %==AVPVC==% \end{equation}
 is an FCI,
 $\motifEx$ is a generalization of a subset of the transactions
 including the transaction $\transaction$
 of equation (\ref{eq:formatage}):
 $\motifEx\subseteq\transaction$.
%This involves that the items of the transaction $\transaction$
% in equation~(\ref{eq:formatage}) frequently appear together
% (with the possible exception of $b\Eg\Pb$).
%The interpretation of this FCI is described hereafter.
%It has been generalized from a set of transactions
% (e.g. the transaction $\transaction$ described in
%  equation~(\ref{eq:formatage})).
The interpretation of this FCI as an adaptation rule is explained
 below.

%             ---------------
\subparagraph{Interpretation.}
%             ---------------
%
The interpretation step is supervised by the analyst.
The \cabamaka system provides the analyst with the extracted FCIs
 and facilities for navigating among them.
%Interpretation is performed by the analyst.
%% (assisted by a computer scientist).
%The \cabamaka system simply assists his/her task by providing
% some means to navigate in the results of the data-mining process.
The analyst may select an FCI, say $\motif$, and interpret $\motif$
 as an adaptation rule.
%The analyst chooses an FCI $\motif$
% and interprets it as an adaptation rule.
For example, the FCI % $\motif$
 $\motifEx$
 %==AVPVC==% in equation~(\ref{eq:exemple-MFF})
 may be interpreted in the following terms:
\begin{quote}
  \strut\llap{\textbf{if~}}\strut
  \num1 $a$ is a property of $\source$ but is not a property of $\cible$, \\
  \num2 $c$ is a property of both $\source$ and $\cible$, \\
  \num3 $d$ is not a property of $\source$ but is a property of $\cible$, \\
  \num4 $A$ and $B$ are properties of $\sol(\source)$ and \\
  \num5 $C$ is not a property of $\sol(\source)$ \\
  % \strut\llap{\textbf{alors}}\strut
  \strut\hspace{-5mm}{\textbf{then}}\strut
  ~the properties of $\sol(\cible)$ are \\
  \strut~~
  $\form(\sol(\cible)) = (\form(\sol(\source))\moins\{A\})\cup\{C\}$.
\end{quote}
This %==AVPVC==% rule
 has to be translated
%==AVPVC==%  from the formalism
%==AVPVC==%  $2^{\Proprietes}$ (sets of boolean properties)
%==AVPVC==%  to the formalism of the adaptation rules of the \rapc system.
%==AVPVC==% The result is
 as an adaptation rule $r$ of the \rapc system.
%==AVPVC==%  \emph{adaptation rule}, i.e. a rule
%==AVPVC==%  whose left part represents
%==AVPVC==%  conditions on $\source$, $\sol(\source)$ and $\cible$
%==AVPVC==%  and whose right part represents
%==AVPVC==%  a way to compute $\sol(\cible)$.
Then the analyst corrects
%==AVPVC==%  and validates
 $r$ and associates an explanation with it.
%==AVPVC==% The analyst is helped in this task by the domain ontology $\Ontologie$
%==AVPVC==%  that is useful to organize the FCIs and by the already available
%==AVPVC==%  adaptation knowledge that is useful to prune from the FCIs the ones
%==AVPVC==%  that are related to already known adaptation knowledge.

%          --------------
\paragraph{Implementation.}
The application domain of the \rapc system we are developing is
 breast cancer treatment:
 in this application, a problem $\prb$ describes a class of
 patients
 with a set of attributes and associated constraints
 (holding on the age of the patient, the size and the localization of
  the tumor, etc.).
A solution $\sol(\prb)$ of $\prb$ is a set of therapeutic decisions
 (in surgery, chemotherapy, etc.).
%==AVPVC==% Two features of this application must be pointed out.
%==AVPVC==% First, the source cases are \emph{general cases}
%==AVPVC==%  (or \emph{ossified cases} according to the terminology
%==AVPVC==%   of~\cite{Riesbeck-Schank89}):
%==AVPVC==%  a source case corresponds to a class of patients and not to a single
%==AVPVC==%  one.
%==AVPVC==% These source cases are obtained from statistical studies in the cancer
%==AVPVC==%  domain.
%==AVPVC==% Second, %and more important,
The requested behavior of the \rapc system
 is to provide a treatment
 and explanations on
 this treatment proposal.
% the construction of the treatment.
This is why the analyst is required to associate an explanation to a
 discovered adaptation rule.

%==AVPVC==% %          ---------------------------------------------------
%==AVPVC==% \paragraph{Representation of cases and of the domain ontology $\Ontologie$.}
%==AVPVC==% %          ---------------------------------------------------
%==AVPVC==% %
The problems, solutions and the domain ontology of the application
 are represented in
%==AVPVC==%  a lite extension of
 OWL~DL
 (recommendation of the W3C).
\section{CONCLUSION}
%        ----------
\label{sec:conclusion}

The \cabamaka system presented in this paper is inspired by the
 research presented in~\cite{hanney96} and by
 the principles of \ecbd for the purpose of semi-automatic adaptation
 knowledge discovery.
%==AVPVC==% It reuses an FCI extraction tool developed in our team
%==AVPVC==%  and based on a \charm-like algorithm.
It has enabled to discover several useful adaptation rules
 for a medical \rapc application.
%==AVPVC==% Although it has been implemented for a specific application
%==AVPVC==%  to breast cancer treatment decision support,
It has been
 designed to be reusable for other \rapc applications:
 only a few modules of \cabamaka are dependent on the formalism
 of the cases and of the domain ontology,
 and this formalism, OWL~DL, is a well-known standard.
One element of future work consists in
 searching for ways of simplifying
 the presentation of the numerous extracted FCIs to the
 analyst.
This involves an organization of these FCIs for the purpose of
 navigation among them.
Such an organization can be a hierarchy of FCIs according to
 their specificities or a clustering of the FCIs in
 themes.

{
  \bibliography{biblio}
}

%%%%%%%%%%%%%%
\end{document}